\documentclass[letterpaper, 10 pt, conference]{ieeeconf}
\IEEEoverridecommandlockouts
\usepackage{xparse}
\usepackage{cite}
\usepackage{amsmath,amssymb,amsfonts}
\usepackage{algorithm2e}
\usepackage{graphicx}
\usepackage{textcomp}
\usepackage{xcolor}
\usepackage{tabularx}
\usepackage{url}
\usepackage{hyperref}

\SetKwInOut{Input}{Input}
\SetKwInOut{Output}{Output}
\SetKwInOut{Parameters}{Parameters}
\SetKwProg{Fn}{def}{:}{}
\SetKwProg{Proc}{procedure}{:}{}
\SetKwProg{Module}{module}{:}{}

\SetCommentSty{mycommentformat}

\SetKwFunction{mabp}{MABP}
\SetKwFunction{astar}{HybridAstar}
\SetKwFunction{pathPlanner}{PathPlanner}
\SetKwFunction{velocityPlanner}{VelocityPlanner}
\SetKwFunction{navPPOH}{NavPPO}
\SetKwFunction{calNavPPO}{calNavPPO}
\SetKwFunction{despot}{IS-DESPOT*}
\SetKwFunction{HyPlan}{HyPlan}
\SetKwFunction{crude}{calibrate}
\SetKwFunction{getConfidence}{getConfidence}

\newcommand{\defined}[1]{\emph{#1}}
\newcommand{\reals}{\mathbb{R}}
\newcommand{\car}{c}
\DeclareDocumentCommand{\exo}{g}{e\IfValueT{#1}{_{#1}}}
\newcommand{\configurations}{\mathcal{C}}
\DeclareDocumentCommand{\state}{g}{s\IfValueT{#1}{^{#1}}}
\DeclareDocumentCommand{\pos}{g}{p\IfValueT{#1}{^{#1}}}
\DeclareDocumentCommand{\goalPos}{g}{\pos{#1}_{\text{goal}}}

\DeclareDocumentCommand{\epos}{g}{\pos{\exo\IfValueT{#1}{_{#1}}}}
\DeclareDocumentCommand{\vel}{g}{v\IfValueT{#1}{^{#1}}}
\DeclareDocumentCommand{\orient}{g}{\theta\IfValueT{#1}{^{#1}}}
\DeclareDocumentCommand{\obs}{g}{o\IfValueT{#1}{^{#1}}}
\DeclareDocumentCommand{\Path}{g}{P\IfValueT{#1}{^{#1}}}

\begin{document}

\title{\LARGE 
HyPlan: Hybrid Learning-Assisted Planning Under Uncertainty\\
for Safe Autonomous Driving
}


\author{Donald Pfaffmann$^{1}$, Matthias Klusch$^{2}$ 
and Marcel Steinmetz$^{3}$ 
\thanks{$^{1}$Saarland University, Computer Science Dept., Saarbrücken, Germany.}
\thanks{$^{2}$German Research Center for Artificial Intelligence (DFKI), Saarbrücken, Germany. 
Contact: {\tt {\small matthias.klusch@dfki.de}}}
\thanks{$^{3}$French National Centre for Scientific Research (CNRS), LAAS-CNRS, Toulouse, France. 
Contact: {\tt {\small marcel.steinmetz@cnrs.fr}}}
}

%
%

\maketitle

\begin{abstract}
We present a novel hybrid learning-assisted planning method, named 
HyPlan, for solving the collision-free navigation problem for 
self-driving cars in partially observable traffic environments. 
HyPlan combines methods for multi-agent behavior prediction, 
deep reinforcement learning with proximal policy optimization 
and approximated online POMDP planning with heuristic 
confidence-based vertical pruning to reduce its execution time 
without compromising safety of driving. 
Our experimental performance analysis on the CARLA-CTS2 benchmark of 
critical traffic scenarios with pedestrians revealed that 
HyPlan may navigate safer than selected relevant baselines and 
perform significantly faster than considered alternative online POMDP planners.
\end{abstract}


\parindent=0pt

\section{Introduction}

\noindent
In general, the collision-free navigation (CFN) problem for
a self-driving car is to minimize the time to a given goal while avoiding
collisions with other objects such as other cars, pedestrians and 
cyclists in a partially observable traffic environment. This 
constrained optimization problem can be modeled as a POMDP and 
solved online by the autonomous vehicle (AV). 
Major classes of the CFN problem solving methods are based on deep learning \cite{cadrl,uniad,vad,kiran21} 
including LLMs \cite{vlad,Cui+24,asyncdriver,llmassist}, 
explicit rule-based \cite{pdmclosedandhybrid} and POMDP action planning \cite{isdespot,cardespot}, 
and hybrid combinations of thereof \cite{leader,hyleap,hylear,pdmclosedandhybrid,letsdrive,lfgnav}.
While hybrid neuro-explicit planning methods such as LEADER \cite{leader}, LFGnav \cite{lfgnav}
and HyLEAP \cite{hyleap} may navigate reasonably safe with provably correct action planning under uncertainty, 
they still suffer from significantly slower execution times compared to deep 
learning-based methods. The main reason is their use of explicit but computationally 
expensive approximate online POMDP planners. One challenge is to close this performance gap 
in terms of execution time and time to goal without compromising safety of driving. 

\noindent
To address this challenge, we developed HyPlan, a novel hybrid learning-assisted planning method 
for collision-free navigation of self-driving cars.
This hybrid method leverages multi-agent behavior prediction, ego-car path planning,
explicit online POMDP planning guided by a PPO-based deep reinforcement learner,
and a recent approach for reducing neural network prediction errors,
in order to obtain approximately optimal car control action policies quickly. 
We conducted a comparative experimental analysis on the CARLA-CTS2 benchmark of critical traffic 
scenes with pedestrian crossings simulated in CARLA \footnote{CARLA: carla.org}, considering
selected state-of-the-art baselines.
Among others, the experimental results revealed that HyPlan may navigate safer 
than all baselines and plan its actions significantly faster than the 
considered alternative explicit and hybrid POMDP planners.
HyPlan is open-source available
\footnote{HyPlan, CARLA-CTS2, baselines: git.opendfki.de/donald.pfaffmann/HyPlan}.

\noindent
The remainder of the paper is structured as follows.
In Section 2, we describe the considered CFN problem as POMDP and 
present our novel hybrid solution HyPlan in Section 3, while the experimental 
evaluation results are summarized in Section 4, before we conclude in Section 5.

\section{Problem Description}

\noindent

The problem of finding control actions for a self-driving car to solve 
the CFN problem in a partially observable traffic environment can be cast 
as a discrete-time POMDP $(S,A,T,R,\gamma,Z,O)$.
The \defined{states} $S = \configurations \times \dots \times \configurations$ describe
the current traffic situation exactly, assuming one single ego-car and $n$ 
exogenous agents. The configuration space of each agents is
$\configurations = \reals^2 \times \reals^2 \times \reals^2 \times [0, 2\pi)$, 
where a state $\state{x}_t = (\pos{x}_t, \goalPos{x}, \vel{x}_t, \orient{x}_t) \in \configurations$ of the agent $x$ at time step $t$ specifies the agent's current and goal positions $\pos{x}_t$ and $\goalPos{x}$, the velocity $\vel{x}_t$, and the orientation $\orient{x}_t$.
For the ego-car, the entire state $\state{\car}_t$ is observable, while for the exo-agents $\exo$, only $\pos{\exo}_t$ is observable, 
if $\exo$ is not occluded.
The \defined{observations} are hence given by $Z = \configurations \times \reals^2 \times \dots \times \reals^2$ with \defined{observation probabilities} $O(s_t, a_t, o_t) = T(s_{t+1} | s_t,a_t)$ for $o = (s^{\car}_{t+1}, p^{\exo{1}}_{t+1}, \dots, p^{\exo{n}}_{t+1})$ (assuming no sensing noise).
We consider car control \defined{actions} $(\alpha, \text{acc}) \in A$ composed of a steering angle $\alpha \in [0, 50]$, and discrete acceleration choice $\text{acc} \in \{\text{Accelerate}, \text{Decelerate}, \text{Maintain}\}$.
The \defined{transition probability function} $T(s_{t+1} | s_t, a_t) \in [0, 1]$ simulates the execution of action $a_t$ in the environment state $s_t$ for a time duration of $\Delta t$, using the bicycle kinematics model \cite{bicycle} to update the car states. Exo-agents are assumed to move to their goal in a straight line.
The \defined{reward function} $R$ is defined so to reward reaching the ego-car's goal as quickly as possible while heavily penalizing collisions and near misses.
The \defined{discount factor} $\gamma = 0.98$ favors near over future rewards.
For more details, we refer to \cite{hylear,despot-cfn}.

\section{Hybrid Solution HyPlan}

\subsection{Overview}

\begin{figure*}[htbp]
\centering
  \includegraphics[width=0.67\textwidth]{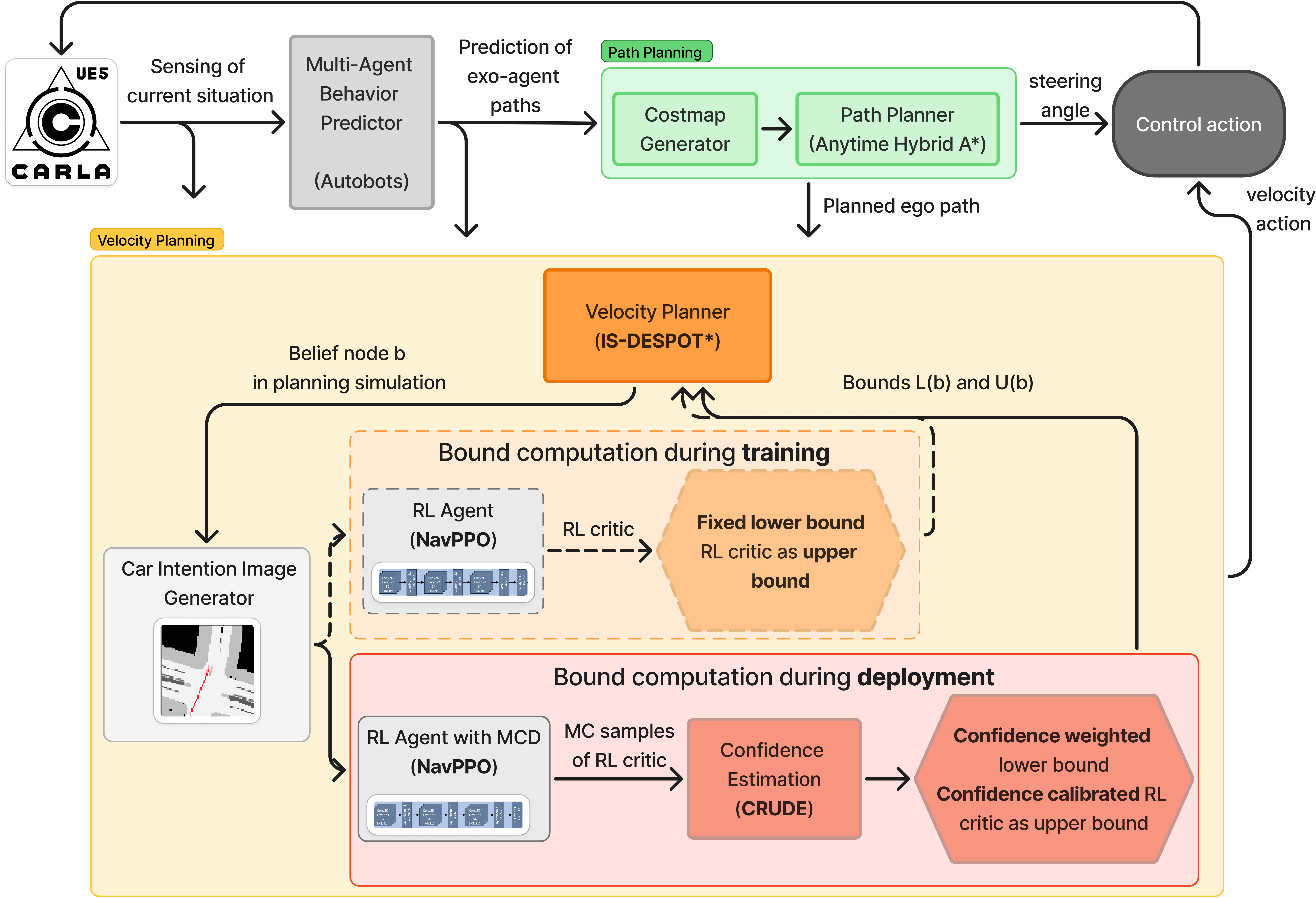}
  \caption{
    Overview of HyPlan system architecture coupled with CARLA. 
     }
 \label{fig:hyplan}
\end{figure*}

HyPlan is a learning-assisted online POMDP planner that solves the above 
mentioned collision-free navigation problem. For this purpose, it integrates
exo-agent trajectory prediction, ego-car path planning, deep reinforcement 
learning with proximal policy optimization, and explicit online POMDP 
planning with heuristic DRL confidence-based vertical pruning to obtain an
approximately optimal control action policy for the ego-car (cf. Fig. 1, Alg. \ref*{alg:hyplan}). 

\begin{algorithm}[htb]
  \small
  \Input{Initial belief $b_0$}
  \Output{Control action pair $(\alpha, \text{acc})$}
  \Parameters{%
      Multi-agent behavior predictor \mabp,\\
      Velocity planner \velocityPlanner
    }

    $c \gets $ car state in $b_0$\;
    $\exo_1^o, \dots, \exo_n^o \gets$ observable states of exo-agents in $b_0$\;
    $\Path^{\exo_1}, \dots, \Path^{\exo_n} \gets$ \mabp{$\exo_1^o, \dots, \exo_n^o$}\;
    $\Path^{\car} \gets \pathPlanner{}$\;
    $\alpha \gets $ steering angle given $c$ and $\Path^{\car}$\;
    $\text{acc} \gets $ \velocityPlanner{}\;
    \Return{$(\alpha, \text{acc})$}\;

    \Module{\pathPlanner{}}{
      $M \gets $ generate costmap representing $c$, $\Path^{\exo_1}, \dots, \Path^{\exo_n}$\;
      $p^{c} \gets $ \astar{$c$, $M$}\;
      \Return{ $p^{\text{c}}$}\;
    }
\vspace*{0.2cm}
  \caption{High-level HyPlan algorithm}
  \label{alg:hyplan}
\end{algorithm}

\noindent

For each scene observation by the ego-car, the MABP AutoBots \cite{autobots} computes exo-agent trajectory predictions. Based on the predictions, a state-of-the-art path planner (weighted hybrid A* \cite{hybrid-astar}) then plans the ego-car path by generating an accordingly enriched 
costmap upon which it computes the shortest and safe path to the goal position.
The next steering action for the ego-car is derived from this 
planned path, while the corresponding best acceleration action for ego-car velocity 
is determined by our new online POMDP planner IS-DESPOT*. 
IS-DESPOT* is a hybrid learning planning method based on the prominent approximate online POMDP action 
planner IS-DESPOT \cite{isdespot}.
Different from IS-DESPOT, IS-DESPOT* leverages an extension of the deep reinforcement learner PPO,
baptized NavPPO, in order to obtain belief
states value estimates and, during deployment, features a novel form of vertical pruning to reduce planning time without impacting
the planning result.
During deployment, HyPlan uses confidence calibration to correct inconsistencies 
in the NavPPO belief state estimates, while its planner IS-DESPOT* leverages 
the confidence estimates to inform the vertical pruning.
During the training phase, the NavPPO network learns to behave as an experience-based 
critic of the IS-DESPOT* planner.
As confidence calibration requires an already trained network, the 
training (cf. Alg. \ref{alg:hyplan-velo-train}, Sect. \ref{training}) 
and deployment (cf. Alg. \ref{alg:hyplan-velo-deploy}, Sect. \ref{testing}) of HyPlan slightly differ.

\subsection{Training of HyPlan}
\label{training}

\noindent
The training modules and procedure of HyPlan are described 
in Alg. \ref*{alg:hyplan-velo-train} and Alg. \ref*{alg:hyplan-train}.
During training, the velocity planning module of HyPlan 
utilizes the hybrid POMDP planner IS-DESPOT* 
to compute the velocity action of the ego-car. The planner requires heuristic functions 
$L$ and $U$ that lower, respectively, upper bound the expected cumulative reward of 
the belief nodes created during the belief tree construction for each time step of 
considered training scene. 
It terminates a planning trial in the belief tree at some belief node $b$, 
if the bound gap $E(b) = L(b) - U(b)$ between the heuristic lower and upper bound 
of its state evaluation is sufficiently small.
For the lower-bound heuristic during training, we fall back to a simple function 
$L_{tr}(b)$ that returns the collision penalty discounted by the number of steps until 
an exo-agent and ego-car moving towards each other collide first.
The heuristic upper bound $U(b)$ is determined by the deep reinforcement 
learner NavPPO. In fact, during the training phase of HyPlan, its NavPPO network 
learns to best estimate the values of given belief states as 
a heuristic upper bound for IS-DESPOT.

\begin{algorithm}[htb]
  \small
  \Parameters{%
    NavPPO critic $V_\theta$,\\
    Lower bound $L_{\text{tr}}$
  }
     \Module{\velocityPlanner{}}{
      \Fn{$L(b_t)$}{\Return{$L_{\text{tr}}(b_t)$}\;}
      \Fn{$U(b_t)$}{\Return{\navPPOH{$b_t$}}\;}
      $\pi^{\text{IS-DESPOT*}} \gets $ \despot{$b_0$} using $L$ and $U$\;
      \Return{sample $\text{acc}$ from $\pi^{\text{IS-DESPOT*}}$}\;
    }
    \Module{\navPPOH{$b_t$}}{
      $I_{b_t} \gets $ generate intention image for $b_t$ and $\Path^{\car}, \Path^{\exo_1}, \dots, \Path^{\exo_n}$\;
      \Return{$V_{\theta}(I_{b_t})$}\;
    } 
\vspace*{0.2cm}
    \caption{HyPlan modules during training}
    \label{alg:hyplan-velo-train}
\end{algorithm}

\begin{algorithm}[htb]
  \small
  \Input{%
    $N$ scenes for training \\
    $M$ scenes for calibration \\
    Number of stochastic forward passes $F$\\
    HyPlan PPO loss function $J_{\text{HyPlan}}$\\
    Max simulation time steps $T_{\text{max}}$ \\
    }
  \Output{NavPPO critic $V_{\theta}$, and\\ CRUDE calibration parameters $\zeta$}
  Initialize actor and critic parameters $\psi$ and $\theta$\;
  \ForEach{training scene}{
    $B \gets \emptyset$\;
    $b_0 \gets $ initial belief of the scene\;
    \ForEach{time step $t=0, \dots, T_{\text{max}}-1$}{
      $a_t \gets $ \HyPlan{$b_t$}\;
      $\pi^{\text{IS-DESPOT*}}_t \gets$ the corresponding IS-DESPOT* policy\;
      $o_t, r_t \gets $ execute $a_t$ in the scene\;
      $b_{t+1} \gets $ update belief according to $o_t$\;
      add $(b_t, \pi^{\text{IS-DESPOT*}}_t, r_t, b_{t+1})$ to $B$\;
    }
    Update $\theta$ and $\psi$ so to minimize $J_{\text{HyPlan}}(\theta, \psi)$ over $B$ via stochastic gradient descent\;
  }
  $E \gets \emptyset$\;
  \ForEach{calibration scene}{
    $B \gets \emptyset$\;
    $b_0 \gets $ initial belief of the scene\;
    \ForEach{time step $t=0, \dots, T_{\text{max}}-1$}{
      $a_t \gets $ \HyPlan{$b_t$}\;
      $o_t, r_t \gets $ execute $a_t$ in the scene\;
      $b_{t+1} \gets $ update belief according to $o_t$\;
      $\mu_t, \sigma^2_t \gets $ statistics of $F$ calls to \navPPOH{$b_t$} with different dropout masks\;
      add $(\mu_t, \sigma^2_t, r_t)$ to $B$\;
    }
    $E \gets E \cup \{(\hat{A}_t - \mu_t) / \sigma_t \mid 0 \leq t < T_{\text{max}},\allowbreak \text{advantage score } \hat{A}_t \text{ in } B \text{ at time } t\}$\; 
  }
  $\zeta \gets $ fit empirical error distribution to $E$\;
  \Return{$(V_{\theta}$, $\zeta)$}\;
  \vspace*{0.2cm}
  \caption{HyPlan training procedure}
  \label{alg:hyplan-train}
\end{algorithm}


\begin{figure*}[thbp!]
\centering
\includegraphics[scale=0.33]{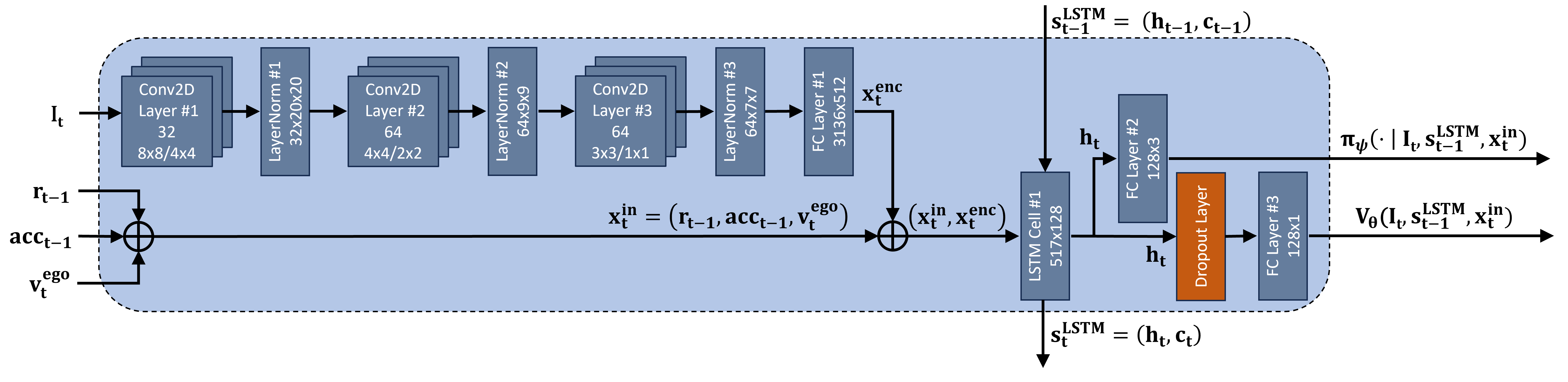}
\caption{DRL NavPPO network architecture} 
 \label{fig:navppo}
\end{figure*}

\noindent
The NavPPO network (cf. Fig. 2) follows a classic actor-critic approach, in which both 
the policy and the value function are modeled via two separate output heads, but share 
a common visual feature encoder. The input to NavPPO is 
(a) an RGB car intention image  $\mathcal{I} \in \mathbb{R}^{84\times 84\times 3}$ 
that enriches the ego-car path planner's costmap with the predicted exo-agent paths, 
the past and the planned path of the car in the scene, and 
(b) non-visual state features 
$\mathbf{x}_t^{\text{in}} = \left(r_{t-1}, \, acc_{t-1}, \, \vel{\car}_t \right)$, 
which are the previously obtained reward, the last executed velocity action and the 
current velocity of the ego-car. 
The visual processing of the car intention image is done by a sequence of three-layered 
convolutional neural networks and normalization followed by fully connected layer for 
final flattening into a latent feature map representation vector that is concatenated 
with the state vector $\mathbf{x}_t^{\text{in}} \in \mathbb{R}^5$ to serve 
together with the LSTM state  $s_{t-1}^{\text{LSTM}}$ from the previous scene step
the final determination of the stochastic action policy 
$\pi_\psi(\cdot \mid I_t, \, s_{t-1}^{\text{LSTM}}, \, x_t^{\text{in}}) 
\in \mathbb{R}^3$ (actor) and the state value estimate 
$V_\theta(I_t, \, s_{t-1}^{\text{LSTM}}, \, x_t^{\text{in}}) \in \mathbb{R}$ 
(critic, in short: $V_\theta(b)$). 

\noindent
In particular, the updated hidden state $h_t$ of the new LSTM state 
$s_t^{\text{LSTM}} = (h_t, c_t)$ branches into two independently trained output paths: 
The upper path realizes the policy function (\textit{Actor}) by means of a fully 
connected layer that maps the 128-dimensional LSTM output onto three neurons, 
representing the logits of a distribution over all possible velocity actions
and thus defining the stochastic policy $\pi_\psi$, while the lower path approximates 
the value function by means of a fully connected layer that maps the LSTM output onto 
a single neuronal unit that provides the scalar value estimate $V_\theta(b)$.
An upstream MC dropout layer is only active during deployment (cf. Section \ref{testing}).

\noindent
NavPPO uses a PPO-based loss function $J(\theta, \psi)$ for its on-policy learning 
of constrained imitation of the planner policy based on the relative quotient of both 
policies $\pi_\psi(acc_t\mid\cdot)$ and $\pi_t^\mathrm{IS-DESPOT*}(acc_t \mid \cdot)$ 
weighed with the estimated benefit of the planned action:
\[J(\theta, \psi) = -J_\pi(\psi) + c J_V(\theta) + \lambda_\text{reg} \left( \|\theta\|_2^2 + \|\psi\|_2^2 \right) \text{, with} \]
\[J_\pi(\psi) = \mathbb{E}_{t \sim \tau}\left[ \min \left( \rho_t(\psi)\hat{A}_t,\; \mathrm{C}(\rho_t(\psi), 1-\epsilon, 1+\epsilon) \hat{A}_t \right) \right] \]
\[\rho_t(\psi) = \frac{\pi_\psi(acc_t \mid I_t, s_{t-1}^\mathrm{LSTM}, x_t^\mathrm{in})}{\pi_t^\mathrm{IS-DESPOT*}(acc_t \mid b_t)} \text{, and } J_V(\theta) = \mathbb{E}_{t \sim \tau} \left[ \hat{A}_t^2 \right]\]
\noindent
where $\theta$ (critic) and $\psi$ (actor) are the network weights to be trained
and updated periodically after each episode, and
$\hat{A}_t$ denotes the generalized advantage estimate (GAE) at time step $t$
\[\hat{A}_t = \sum_{l=0}^{T - t - 1} (\gamma \lambda)^l \delta_{t + l} \text{, with temporal-difference residual }\]
\[ \delta_t = r_t + \gamma V_\theta(I_{t+1}, s_t^\mathrm{LSTM}, x_{t+1}^\mathrm{in}) - V_\theta(I_t, s_{t-1}^\mathrm{LSTM}, x_t^\mathrm{in})\]
where $\lambda \in [0, 1]$ is a decay factor.
The function $\rho_t(\psi)$ denotes the probability ratio between the current DRL policy 
$\pi_{\psi}$ and the policy $\pi_t^\mathrm{IS-DESPOT*}(acc_t \mid b_t)$ proposed by 
the planner for executing action $acc_t$ at scene simulation step $t$ under belief $b_t$.
$\mathrm{C}(\cdot)$ is the clip operator that restricts $\rho_t(\psi)$ to the interval
$[1 - \epsilon, \, 1 + \epsilon]$ to prevent excessively large policy updates,
$\epsilon \in \mathbb{R}_{>0}$ the clipping threshold, 
$c \in \mathbb{R}_{\geq 0}$ the weighting factor controlling the contribution of 
the value loss, and 
$\lambda_\text{reg} \in \mathbb{R}_{\geq 0}$ the regularization coefficient jointly 
applied to all actor-critic network parameters to prevent overfitting.

\noindent
After NavPPO has been successfully trained, the NavPPO value predictions are validated 
over a separate calibration instance set.
To this end, HyPlan is simulated over the calibration scenes, collecting during this 
process the mean and standard deviations of $F$ stochastic forward passes of the NavPPO 
value network with Monte Carlo dropouts for each simulation step; 
in our experiments, we set $F$ = 10. Based on the collected statistics, normed empirical 
errors with respect to the GAE are computed. At the end, an empirical distribution for 
the collected normalized errors is constructed, which during the deployment of HyPlan, 
serves as input to CRUDE \cite{crude} to calibrate the NavPPO value estimates and the 
confidence in them.


\subsection{Deployment Architecture of HyPlan}
\label{testing}

\begin{algorithm}[htb]
  \Parameters{%
    NavPPO critic $V_\theta$,\\
    Manually designed lower bound $L_{\text{tr}}$,\\
    CRUDE calibration parameters $\zeta$,\\
    Number of stochastic forward passes $F$
  }
  \small
     \Module{\velocityPlanner{}}{
      \Fn{$L(b_t)$}{
        $\tilde{\mu}, \tilde{\sigma}^2 \gets $ \calNavPPO{$b_t$}\;
     $\varphi = $ \getConfidence{$\tilde{\sigma}^2$}\;
     \Return{$(1 - \varphi) L_{\text{tr}}(b_t) + \varphi \tilde{\mu}$}\;
      }
      \Fn{$U(b_t)$}{
        $\tilde{\mu}, \tilde{\sigma}^2 \gets $ \calNavPPO{$b_t$}\;
        \Return{$\tilde{\mu}$}\;
      }
      $\pi^{\text{DESPOT}} \gets $ \despot{$b_0$} using $L$ and $U$\;
      \Return{sample $\text{acc}$ from $\pi^{\text{DESPOT}}$}\;
    }

    \Module{\calNavPPO{$b_t$}}{
      $I_{b_t} \gets $ generate intention image for ${b_t}$ and $\Path^{\car}, \Path^{\exo_1}, \dots, \Path^{\exo_n}$\;
      $\mu, \sigma^2 \gets $ $F$ samples of $V_{\tilde{\theta}}(I_{b_t})$ using dropout\;
      $\tilde{\mu}, \tilde{\sigma}^2 \gets $ \crude{$\mu$, $\sigma^2$; $\zeta$}\;
     \Return{$\tilde{\mu}, \tilde{\sigma}^2$}\;
    } 
\vspace*{0.2cm}
    \caption{HyPlan modules during deployment}
    \label{alg:hyplan-velo-deploy}
\end{algorithm}

\noindent
In the deployment phase of HyPlan, the online POMDP planner IS-DESPOT* 
additionally performs vertical pruning of its action planning based on the confidence 
the trained learner NavPPO has in its experience-based evaluation of belief states. 
More concretely, NavPPO runs multiple stochastic forward passes with different dropout 
masks to obtain the mean $\mu$ of generated belief state value estimates 
$V_\theta(b)$ together with their variance $\sigma^2$. 
The planner receives from the learner the calibrated mean $\tilde\mu$ 
of state values as heuristic upper bound $U(b)$ and the calibrated variance 
$\tilde\sigma^2$, and uses the inverse of the latter as 
the confidence value $\varphi \in [0,1]$ of the learner in its evaluation of belief state $b$. 
The calibrations are performed with the standard method CRUDE \cite{crude}, 
in particular to reduce cases of false positives\footnote{For example, in such cases, 
NavPPO would return a high-confidence high-value estimate for a belief node which plan branch 
after node selection would later on lead to a collision, even if the car were to continuously 
decelerate in this branch.}.

\noindent
For each belief node $b$ created during planning, both its state value estimation 
and the confidence in it by the learner are used by the planner to determine whether 
it is worthwhile to continue planning from $b$ in the belief tree, or rather not. 
For this purpose, the planner uses the heuristic lower bound $L(b)$, which 
is defined as confidence-weighted sum $L(b) = (1-\varphi_b)L_{tr}(b) + \varphi_b U(b)$ of 
the pre-defined heuristic lower bound $L_{tr}(b)$ as utilized during training 
(cf. Sect. \ref{training}) and the heuristic upper bound $U(b)$ from the trained learner.
Therefore, highly confident upper bound value estimations for selected belief states 
result in smaller bound gaps than those with lower confidence. 

\noindent
In particular, belief states with maximal upper bound value estimate and high confidence 
value by the learner lead to an earlier termination of the planning trial and final action 
selection at the root node than without this kind of vertical pruning. That heuristically 
avoids over-exploration of familiar regions of the belief state space by the planner for 
which the state value estimations returned by the learner are unlikely to change, which 
is indicated by the degree of confidence the learner has in them. 

\noindent
As a simple example of vertical pruning, consider some traffic scene simulation step 
for which enriched observation IS-DESPOT* starts its planning of an approximately optimal 
velocity action for the ego-car. Let us further assume that no collision threat 
would be left it accelerates at this step, such that the car could even continue 
accelerating until the goal is reached without any collision and NavPPO learned to recognize 
and evaluate such situations with high confidence. In this example, the planner reaches 
already on the first planning layer a belief state for simulated action "Accelerate" 
from the root which gets evaluated with the maximal upper bound value estimate 
and with high confidence by NavPPO. As a consequence, the planner terminates the 
whole planning trial, selects this action after the backup process to the root 
as heuristically optimal for this scene step, and executes it. In particular, 
the final selection of this action would not have changed in subsequent planning steps 
in this trial that would be needed without pruning. 
One question is whether and to what extent the potential speed-up of online POMDP 
planning through this learner confidence-based vertical pruning would come at the cost 
of safety of driving compared to other baseline methods in general.

\section{Evaluation}

\noindent
We conducted an experimental comparative performance evaluation of HyPlan 
with selected baselines regarding safety, efficiency and planning efforts 
on the synthetic CARLA-CTS2 benchmark of critical traffic scenarios.

\subsection{Experimental setting}

\noindent
{\em Benchmark.} The CARLA-CTS2 benchmark consists of nine parameterized 
scenarios with about twenty-three thousand traffic scenes with pedestrian 
crossing simulated with the CARLA driving simulator.
These traffic scenarios are identified in the real-world in-depth accident 
study in Germany (GIDAS) \cite{GIDAS} as most critical, where the car is 
confronted with a street crossing pedestrian, possibly occluded by some parking car, 
an incoming car, and different street intersections. The scenes per 
scenario are generated with varying speed and crossing distance of 
pedestrians from the car.   

\begin{figure*}[thbp]
\centering
\includegraphics[scale=0.6]{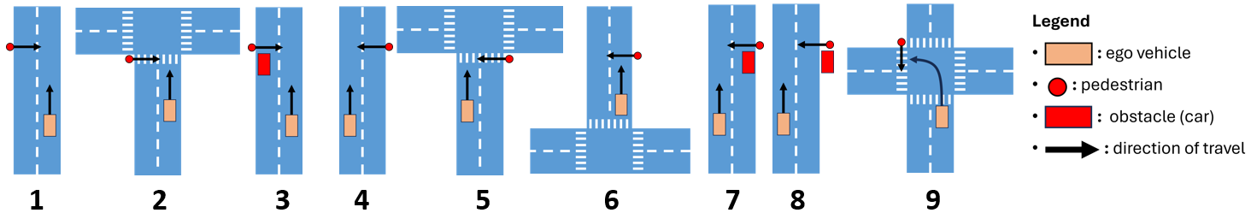}
\caption{Scenarios of CARLA-CTS2 benchmark with pedestrian crossing} 
 \label{fig:carla-cts}
\end{figure*}


\noindent
{\em Baselines.} The selected baseline methods for CFN problem solving are 
(a) the deep reinforcement learners NavPPO and NavA2C,
(b) the explicit navigation planner IS-DESPOTp \cite{isdespot},
(c) the hybrid learning-assisted planners HyLEAP \cite{hyleap} 
    and LEADER \cite{leader}, and
(d) the hybrid planning-assisted deep reinforcement learner HyLEAR \cite{hylear}.
Training, calibration and testing on the benchmark followed
a 25:25:50\% ratio of respective sets.
The explicit navigation planner IS-DESPOTp utilizes the IS-DESPOT 
planner \cite{isdespot} to obtain the approximately optimal velocity 
for actual steering action derived from ego-car path planned with 
standard hybrid A* used in all baselines.

\noindent
{\em Measurements.}
The performance of each method is measured in terms of 
(a) the overall safety index (SI90) defined as total number of
    scenarios in which the method fails (crash or near-miss)
    in 10 percentages of the scenes;
(b) the crash and near-miss rates (\%), and time to goal (TTG) 
    in seconds; and 
(c) the execution time in milliseconds and, if applicable,
    training time in days.
In addition, for explicit planning and hybrid methods, the 
effort of planning is measured in terms of average
(a) planning time (PT), (b) number of planning trials (PTN),
(c) planning trial depth (PTD), (d) number of belief nodes in BT (BNN),
(e) observation branching factor (OBF), and (f) neural network 
evaluation time (NNET).  
HyPlan is implemented in Python and PyTorch framework;
all experiments were run on the NVIDIA DGX-2 servers.

\subsection{Results}

\begin{table*}[htbp]
    \centering
    \begin{tabular}{|c|c|c|c|c|c|c|}
        \hline
   \textbf{Method} & ~Safety~ & ~Crash (\%)~ & ~Near-Miss (\%)~ & ~TTG (s)~ & ~Execution (ms)~ & ~Training (d)~\\
         \hline
     IS-DESPOTp & 1 & 18.43 & 18.94 & 16.91 & 251.29 & N/A  \\
         \hline
     NavA2C     & 1 & 13.50 & 24.02 & 13.74 & 104.95 & 2.51 \\
         \hline
     NavPPO     & 2 & 14.64 & 10.24 & {\bf 12.74} & {\bf 32.59} & {\bf 1.91} \\
         \hline
     HyLEAR     & 3 & 13.76 & 14.15 & 15.21 & 41.38 & 3.21  \\
         \hline     
     HyLEAP     & 4 & 13.29 & 13.39 & 16.46 & 225.76 & 4.76 \\
         \hline
     LEADER     & 4 & 12.64 & 16.67 & 16.72 & 267.43 & 5.01 \\
         \hline
     HyPlan     & {\bf 7} & {\bf 12.36} & {\bf 8.04} & 19.97 & 161.17 & 4.33 \\
         \hline
    \end{tabular}
    \vspace*{0.2cm}
    \caption{Experimental results for navigation safety and efficiency of HyPlan and  
             baselines over benchmark CARLA-CTS2}
    \label{tab:performance_overview}
\end{table*}

\noindent
The overall results of our experiments with respect to safety, 
efficiency, and, where applicable, the planning efforts of 
the considered methods, averaged across all scenarios, are shown in 
Tables I and II. For each measure, we first average the scores across all 
scenes of a scenario and then average across all scenarios such that each 
scenario is weighted equally.

\noindent
In general, HyPlan provided a safer ride than all baselines (cf. Table 
I). Moreover, its execution is significantly faster (up to 40 percent) 
than all other explicit and hybrid planning baselines, but still
remains three to four times slower than the considered learning 
baselines and at the expense of the highest TTG value
due to a more cautious driving behavior. 
The deep reinforcement learner NavPPO is fastest in training, 
execution and with best TTG value, but that comes at the very cost of 
safety. Overall, navigation with the considered hybrid methods turned 
out to be safer than with the tested methods of explicit planning
and mere deep learning. 

\noindent
Regarding the planning efforts (cf. Table II), HyPlan turned out 
to incorporate the fastest apprixmate online POMDP planner
(IS-DESPOT with learner confidence-based vertical pruning), 
with shortest planning trial depths and less belief nodes to evaluate 
in less planning trials on average, compared to all alternative 
planners. The hybrid learning-assisted planner LEADER was slowest 
in planning and execution but fastest in its used neural network 
evaluation of the situation as input to but not during planning 
in IS-DESPOT.  

\begin{table*}[htbp]
    \centering
    \begin{tabular}{|c|c|c|c|c|c|c|}
        \hline
   \textbf{Method} & ~PT (ms)~ & ~NNET (ms)~ & ~BNN~ & ~PTN~ & ~PTD~ & ~OBF~\\
         \hline
    IS-DESPOTp & 244.72 & N/A & 543 & 43.91 & 9.29 & 1.23 \\
         \hline
    HyLEAP     & 218.71 & 187.93 & 39.54 & 2.36 & 6.46 & 1.15 \\
         \hline
    LEADER     &  259.58 & {\bf 6.76} & 568 & 45.73 & 9.47 & 1.32 \\
         \hline
    HyPlan     &  {\bf 149.94} & 116.52 & {\bf 15.62} & 3.62 & {\bf 2.21} & {\bf 0.93} \\
         \hline
    \end{tabular}
    \vspace*{0.2cm}
    \caption{Experimental results for planning efforts of explicit and hybrid planners 
             over CARLA-CTS2.\\
    PT: Planning Time; BNN: \#Belief nodes; PTN: \#Planning Trials; 
    PTD: Planning Trial Depth; OBF: Observation Branching Factor; 
    NNET: Neural Network Evaluation Time.}
    \label{tab:performance_overview}
\end{table*}


\noindent
In addition, we conducted an extensive ablative study of HyPlan to
investigate the impact of its individual functional components for 
path prediction, DRL, confidence calibration and vertical pruning 
in the planning process on the overall performance of the system 
Among others, the results showed that compared to the full HyPlan, 
the best navigation safety can be achieved with its variant that 
avoids vertical pruning while calibrating its belief node 
evaluations for the planner, though at the cost of longest 
execution time out of all variants. On the other hand, avoiding 
confidence calibration resulted in over-optimistic (confidence for) 
vertical pruning of the planner, which led to the fastest execution 
time of all variants but at the cost of worst navigation safety. 
Interestingly, vertical pruning alone caused the respective variant 
of HyPlan to be significantly faster than all its other variants 
with planning but showed slightly less safe driving behavior than 
the variants without vertical pruning. The additional integration 
of pedestrian path prediction (and its use for planning) led to 
marginally increased navigation safety and reduced planning efforts 
but at the cost of slightly increased execution time. 
Finally, the combination of all new or improved methods in HyPlan 
compared to its variants led to the best tradeoff with respect to 
safety of driving and execution time.

\section{Related Work}

\noindent
As mentioned above, there exists a wide range of methods to address 
the problem of collision-free navigation under uncertainty.
Deep learning-based approaches have become omnipresent in autonomous driving 
research \cite{grigorescu-etal-2020,kiran21}. These approaches span from learning vehicle control 
policies \cite{kendall-etal-2019}, e.g., via deep reinforcement learning, operating on semantic 
environment features, to end-to-end learning approaches \cite{Chen+24}, such as UniAD \cite{uad}, 
that consolidate the entire autonomous driving decision pipeline into a single deep neural network.
Sparked by the success of LLMs in other decision-making tasks, the use of LLMs for 
autonomous driving is receiving significant attention lately such as 
\cite{llmassist,lmdrive,drivevlm,emma,asyncdriver,pdmclosedandhybrid} 
with noticeable success on relevant benchmarks such as nuPlan \cite{nuplan}. 
For example, LMdrive \cite{lmdrive} leverages a multi-modal LLM, which fuses language instructions 
with camera and LiDAR sensor data to low-level control outputs. 
However, since the decision-making process is entirely based on black-box models, these
end-to-end learning approaches suffer from severe predictability and robustness issues 
\cite{Chen+24}. By combining experience-based with explicit planning approaches, those issues are 
much less pertinent in HyPlan. In general, the use of LLMs as planners without any 
integrated logic-based verification as safeguard \cite{veriplan} does not yield logically 
provable correct planned policies with guarantees \cite{llmasplannersurvey}. 

\noindent
Alternative neuro-explicit planning approaches are by no means new and have been explored 
in various works in the past \cite{leader,hyleap,lfgnav,letsdrive}.
At a high level, one can classify these hybrid approaches into learning-assisted planners and 
planning-assisted learners. As an example for planning-assisted learners, HyLEAR \cite{hylear} 
uses an explicit planner to improve training of a vehicle behavior-control policy that optimizes 
simultaneously the satisfaction of passenger preferences and traffic rules as well as avoiding 
collisions. More relevant to HyPlan is the category of learning-assisted explicit planners. 
For instance, LEADER \cite{leader} learns a model predicting pedestrian intentions, which is used 
to steer the Monte Carlo sampling in their explicit IS-DESPOT planner to critical situations.
This is orthogonal to HyPlan, which leverages an external behavior predictor to construct more 
informed inputs for an explicit path planner and the PPO-based belief node estimator.
LFG \cite{lfgnav} uses common knowledge embedded into LLMs in order to implement a search guidance 
heuristic for a navigation planner that takes a natural language goal. 
Similarly, HyPlan leverages an experience-based search guidance heuristic.
However, the heuristic in HyPlan is not designed to support natural language goals and is 
optimized directly for the purpose of guiding its explicit POMDP planner IS-DESPOT 
through the learner NavPPO imitating the planner policy. 


\section{Conclusion}

We presented HyPlan, a novel hybrid learning-assisted online POMDP planner 
for collision-free driving control action policies for autonomous cars. 
The experimental results over the CARLA-CTS benchmark revealed that HyPlan can 
outperform all selected baselines in terms of safety. Moreover, it is significantly 
faster than all selected explicit planning-based methods in training and execution time,
and further reduces but does not yet close the existing inference speed gap 
to the considered mere deep learning-based CFN methods. 
Ongoing work is concerned with evaluating HyPlan over an even broader range 
of traffic scenarios such as in nuPlan \cite{nuplan}, and 
improving the integrated explicit planning process through its use of 
additional scene information. \\

\noindent
{\bf Acknowledgement.} 
This work has been funded by the German Ministry for Research, Technology and Space (BMFTR) in project Momentum, and the European Commission in project InnovAIte.


\begin{thebibliography}{99}

\bibitem{despot-cfn}
Bai, H., Cai, S., et al. (2015).
Intention-aware online POMDP planning for autonomous driving in a crowd.
Proc. International Conference on Robotics and Automation (ICRA). IEEE.

\bibitem{GIDAS}
Bartels, B. \& Liers, H. (2014):
Bewegungsverhalten von Fussgaengern im Strassenverkehr, Teil 2.
FAT-Schriftenreihe, Nr. 268.

\bibitem{letsdrive}
Cai, P., \& Hsu, D. (2022). 
Closing the planning–learning loop with application to autonomous driving. 
{\em IEEE Transactions on Robotics}, 39(2). IEEE.

\bibitem{cardespot}
Cannizzaro, R., \& Kunze, L. (2023). 
Car-DESPOT: Causally-informed online pomdp planning for robots in confounded 
environments. 
Proc. IEEE/RSJ International Conference on Intelligent Robots and Systems (IROS). IEEE.

\bibitem{llmasplannersurvey}
Cao, P., Men, T., et al. (2025). 
Large language models for planning: A comprehensive and systematic survey. 
arXiv preprint arXiv:2505.19683.

\bibitem{asyncdriver}
Chen, Y., Ding, Z.H., et al. (2024). 
Asynchronous large language model enhanced planner for autonomous driving. 
Proc. European Conference on Computer Vision (ECCV). Springer. 

\bibitem{Chen+24}
Chen, L., Wu, P., et al. (2024). 
End-to-end autonomous driving: Challenges and frontiers. 
{\em IEEE Transactions on Pattern Analysis and Machine Intelligence}. 46(12). IEEE.

\bibitem{Cui+24}
Cui, C., et al. (2024). 
A survey on multimodal large language models for autonomous driving. 
Proc. IEEE/CVF Winter Conference on Applications of Computer Vision.

\bibitem{leader}
Danesh, M.H., Cai, P., \& Hsu, D. (2023). 
LEADER: Learning attention over driving behaviors for planning under uncertainty. 
Proc. International Conference on Robot Learning (CoRL). PMLR.

\bibitem{pdmclosedandhybrid}
Dauner, D., et al. (2023). 
Parting with misconceptions about learning-based vehicle motion planning. 
Proc. of Machine Learning Research (PMLR) for Conference on Robot Learning (CoRL).
See also: Dauner, D., et al. (2023). Predictive Driver Model: A Technical Report.

\bibitem{hybrid-astar}
Dolgov, D., Thrun, S., Montemerlo, M., \& Diebel, J. (2010).
Path planning for autonomous vehicles in unknown semi-structured environments. 
{\em Robotics Research}, 29(5).

\bibitem{cadrl}
Everett, M., Chen, Y.F., \& How, J.P. (2021): 
Collision avoidance in pedestrian-rich environments with deep reinforcement learning. 
{\em IEEE Access}, 9:10357–10377. IEEE.

\bibitem{vlad}
Gariboldi, C., Tokida, H., Kinjo, K., Asada, Y., \& Carballo, A. (2025). 
VLAD: A VLM-Augmented Autonomous Driving Framework with Hierarchical Planning 
and Interpretable Decision Process. 
arXiv preprint arXiv:2507.01284.

\bibitem{autobots}
Girgis, R., et al. (2021). 
Latent variable sequential set transformers for joint multi-agent motion prediction. 
arXiv preprint arXiv:2104.00563.

\bibitem{grigorescu-etal-2020}
Grigorescu, S., Trasnea, B., Cocias, T., \& Macesanu, G. (2020).
A survey of deep learning techniques for autonomous driving.
{\em Field Robotics}, 37(3).

\bibitem{uad}
Guo, M., Zhang, Z., He, Y., Wang, K., \& Jing, L. (2024). 
End-to-end autonomous driving without costly modularization and 3d manual annotation. 
arXiv preprint arXiv:2406.17680.

\bibitem{hylear}
Gupta, D. \& Klusch, M. (2023): 
Hybrid deep reinforcement learning and planning for safe and comfortable 
automated driving.  
Proc. 34th IEEE International Intelligent Vehicles Symposium (IV), IEEE.  

\bibitem{uniad}
Hu, Y., et al. (2023). 
Planning-oriented autonomous driving. 
Proceedings of the IEEE Conference on Computer Vision and Pattern Recognition 
(CVPR). IEEE.

\bibitem{emma}
Hwang, J. J., Xu, R., et al. (2024).
Emma: End-to-end multimodal model for autonomous driving.
arXiv preprint arXiv:2410.23262.

\bibitem{vad}
Jiang, B., Chen, S., Xu, Q., Liao, B., Chen, J., Zhou, H., ... \& Wang, X. (2023). 
VAD: Vectorized scene representation for efficient autonomous driving.
Proc. IEEE/CVF International Conference on Computer Vision.

\bibitem{kendall-etal-2019}
Kendall, A., Hawke, J., et al. (2019).
Learning to drive in a day. 
Proc. IEEE International Conference on Robotics and Automation (ICRA). IEEE.

\bibitem{kiran21}
Kiran, B.R., et al. (2021). 
Deep reinforcement learning for autonomous driving: A survey. 
{\em IEEE Transactions on Intelligent Transportation Systems}, 23(6). IEEE.

\bibitem{bicycle}
Kong, J., Pfeiffer, M., Schildbach, G., \& Borrelli, F. (2015).
Kinematic and dynamic vehicle models for autonomous driving control design. 
Proc. IEEE Intelligent Vehicles Symposium. IEEE.

\bibitem{veriplan}
Lee, C.P., Porfirio, D., et al. (2025). 
Veriplan: Integrating formal verification and llms into end-user planning. 
Proc. CHI Conference on Human Factors in Computing Systems.

\bibitem{isdespot}
Luo, Y., et al. (2019): 
Importance sampling for online planning under uncertainty. 
{\em Robotics Research}, 38(2-3).

\bibitem{nuplan}
nuPlan Benchmark for autonomous vehicle planning.
https://www.nuscenes.org/nuplan

\bibitem{hyleap}
Pusse, F., \& Klusch, M. (2019): 
Hybrid online POMDP planning and deep reinforcement learning for safer 
self-driving cars. 
Proc. IEEE Intelligent Vehicles Symposium (IV). IEEE.

\bibitem{lfgnav}
Shah, D. et al. (2023). 
Navigation with large language models: Semantic guesswork as a heuristic for 
planning. 
Proc. of Machine Learning Research (PMLR) for Conference on Robot Learning.

\bibitem{lmdrive}
Shao, H., Hu, Y., et al. (2024).
Lmdrive: Closed-loop end-to-end driving with large language models.
Proc. {IEEE/CVF} Conference on Computer Vision and Pattern Recognition.

\bibitem{llmassist}
Sharan, S.P., Pittaluga, F., \& Chandraker, M. (2023). 
Llm-assist: Enhancing closed-loop planning with language-based reasoning. 
arXiv preprint arXiv:2401.00125.

\bibitem{drivevlm}
Tian, X., Gu, J., et al. (2024).
Drivevlm: The convergence of autonomous driving and large vision-language models.
arXiv preprint arXiv:2402.12289.

\bibitem{crude}
Zelikman, E., et al. (2020). 
CRUDE: calibrating regression uncertainty distributions empirically. 
arXiv preprint arXiv:2005.12496.














\end{thebibliography}

\end{document}